\documentclass[10pt,twocolumn,letterpaper]{article}

\usepackage{iccv}
\usepackage{times}
\usepackage{epsfig}
\usepackage{graphicx}
\usepackage{amsmath}
\usepackage{amssymb}

\usepackage{algorithm}
\usepackage{algorithmic}
\usepackage{mathrsfs}

\usepackage{graphicx}
\usepackage{times}
\usepackage{epsfig}
\usepackage{graphicx}
\usepackage{amsmath}
\usepackage{amssymb}
\usepackage{xcolor}
\usepackage{caption}
\usepackage{subcaption}
\usepackage{tabularx}
\usepackage{rotating}
\usepackage{verbatim}
\usepackage{xcolor,colortbl}
\usepackage{etoolbox}
\usepackage[normalem]{ulem}
\usepackage{booktabs}
\usepackage{multirow}
\usepackage{lipsum}
\usepackage{stmaryrd}
\usepackage{stackengine}
\usepackage{makecell}
\usepackage{pifont}
\usepackage{cancel}
\usepackage{adjustbox}
\usepackage{dblfloatfix}
\usepackage{bbding}
\usepackage{pifont}
\usepackage{wasysym}
\usepackage{amssymb}
\usepackage[pagebackref=true,breaklinks=true,letterpaper=true,colorlinks,bookmarks=false]{hyperref}

\definecolor{barrier}{RGB}{112,128,144}
\definecolor{bicycle}{RGB}{220,20,60}
\definecolor{bus}{RGB}{255, 127, 80}
\definecolor{car}{RGB}{255, 158, 0}
\definecolor{const. veh.}{RGB}{233, 150, 70}
\definecolor{motorcycle}{RGB}{255,61,99}
\definecolor{pedestrian}{RGB}{0,0,230}
\definecolor{traffic cone}{RGB}{47,79,79}
\definecolor{trailer}{RGB}{255,140,0}
\definecolor{truck}{RGB}{255,99,71}
\definecolor{drive. suf.}{RGB}{0,207,191}
\definecolor{other flat}{RGB}{175,0,75}
\definecolor{sidewalk}{RGB}{75,0,75}
\definecolor{terrain}{RGB}{112,180,60}
\definecolor{manmade}{RGB}{222,184,135}
\definecolor{vegetation}{RGB}{0,175,0}

\newcommand{\cmark}{\ding{51}}%
\newcommand{\xmark}{\ding{55}}%
\iccvfinalcopy % *** Uncomment this line for the final submission
\def\iccvPaperID{6588} % *** Enter the ICCV Paper ID here

\ificcvfinal\fi

\begin{document}

%%%%%%%%% TITLE
\title{OpenOccupancy: A Large Scale Benchmark for \\Surrounding Semantic Occupancy Perception}
\author{
Xiaofeng Wang\textsuperscript{\rm 1,3}\footnotemark[1]~~Zheng Zhu\textsuperscript{\rm 2}\footnotemark[1]~\footnotemark[2]~~Wenbo Xu\textsuperscript{\rm 2}\footnotemark[1]~~Yunpeng Zhang\textsuperscript{\rm 2}\\
Yi Wei\textsuperscript{\rm 4}~~Xu Chi\textsuperscript{\rm 2}~~Yun Ye\textsuperscript{\rm 2}~~Dalong Du\textsuperscript{\rm 2}~~Jiwen Lu\textsuperscript{\rm 4}~~Xingang Wang\textsuperscript{\rm 1}$^{\dagger}$\\
\textsuperscript{\rm 1}Institute of Automation, Chinese Academy of Sciences
~ ~ \textsuperscript{\rm 2}PhiGent Robotics \\
~ ~ \textsuperscript{\rm 3}University of Chinese Academy of Sciences
~ ~ \textsuperscript{\rm 4}Tsinghua University \\
%\tt\small \{wangxiaofeng2020,xingang.wang\}@ia.ac.cn
%zhengzhu@ieee.org lujiwen@tsinghua.edu.cn\\
%\tt\small \{wenbo.xu,yunpeng.zhang,xu.chi,yun.ye\}@phigent.ai 
%y-wei19@mails.tsinghua.edu.cn
}

\twocolumn[{%
\vspace{-1em}
\maketitle
\vspace{-1em}

\begin{center}
\vspace{-2em}
\centering
\resizebox{1\linewidth}{!}{
\includegraphics{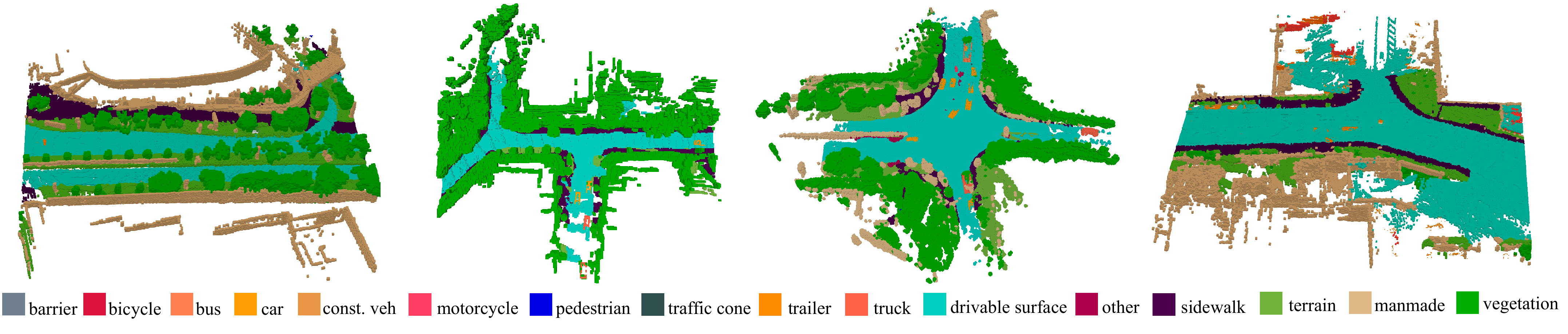}}
\captionof{figure}{The nuScenes-Occupancy provides dense semantic occupancy labels for all key frames in the nuScenes \cite{nusc} dataset. Here we showcase the annotated ground truth with the volumetric size of $(40\times 512\times 512)$ and grid size of 0.2~m.}
\label{fig:showgt}
\end{center}}] 

\renewcommand{\thefootnote}{\fnsymbol{footnote}} %将脚注符号设置为fnsymbol类型，即特殊符号表示
\footnotetext[1]{These authors contributed equally to this work.} %对应脚注[1]
\footnotetext[2]{Corresponding authors. zhengzhu@ieee.org, xingang.wang@ia.ac.cn} %对应脚注[2]
\footnotetext[3]{\url{https://github.com/JeffWang987/OpenOccupancy}}
\ificcvfinal\thispagestyle{empty}\fi

%%%%%%%%% ABSTRACT
\begin{abstract}
\vspace{-1em}

Semantic occupancy perception is essential for autonomous driving, as automated vehicles require a fine-grained perception of the 3D urban structures. However, existing relevant benchmarks lack diversity in urban scenes, and they only evaluate front-view predictions. Towards a comprehensive benchmarking of surrounding perception algorithms, we propose OpenOccupancy, which is the first surrounding semantic occupancy perception benchmark. In the OpenOccupancy benchmark, we extend the large-scale nuScenes dataset with dense semantic occupancy annotations. Previous annotations rely on LiDAR points superimposition, where some
occupancy labels are missed due to sparse LiDAR channels. To mitigate the problem, we introduce the \textbf{A}ugmenting \textbf{A}nd \textbf{P}urifying (AAP) pipeline to $\sim$2$\times$ densify the annotations, where $\sim$4000 human hours are involved in the labeling process.
Besides, camera-based, LiDAR-based and multi-modal baselines are established for the OpenOccupancy benchmark. Furthermore, considering the complexity of surrounding occupancy perception lies in the computational burden of high-resolution 3D predictions, we propose the Cascade Occupancy Network (CONet) to refine the coarse prediction, which relatively enhances the performance by $\sim$30\% than the baseline. 
We hope the OpenOccupancy benchmark \footnotemark[3] will boost the development of surrounding occupancy perception algorithms.
\looseness=-1

\end{abstract}

\begin{table*}[t]
\centering
\small{
\begin{tabular}{lcccccccc}
\toprule
 &Type& Surround & Modality&Vol. Size & \#Scenes & \#Frames & Annotation   \\
\midrule
NYUv2~\cite{nyu} &Indoor& \xmark & C\&D  & $(144\times 240 \times240)$ & 1.4K & 1.4K & Human   \\
ScanNet~\cite{scannet} &Indoor& \xmark& C\&D & $(31\times 62 \times 62)$ & 1.5K & 1.5K & Human \\
SceneNN~\cite{scenenn} &Indoor& \xmark & C\&D & - & 100 & - & Human  \\
SUNCG~\cite{suncg} &Synthtic& \xmark & C\&D  & $(144\times 240 \times240)$ & 46K & 140K & Synthtic   \\
 SynthCity~\cite{synthcity} &Synthtic& \xmark & L & - & 9 & - & Synthtic \\
 SemanticPOSS~\cite{semanticposs} &Outdoor& \cmark & L & - & - & 3K & Human  \\
 SemanticKITTI~\cite{semantickitti}  &Outdoor& \xmark & C\&L & $(32\times 256 \times 256)$ & 22 & 9K & Human   \\
\midrule
\textbf{nuScenes-Occupancy} &Outdoor&  \cmark &  C\&L & $(40\times 512 \times 512)$ & 850  & 200K$^1$ & Auto\&Human \\
\bottomrule
\vspace{-0.5cm}
\end{tabular}}
\caption{Comparison between nuScenes-Occupancy and other dense LiDAR/occupancy perception datasets. \textit{Surround=\cmark} represents datasets that use surround-view inputs. \textit{C, D, L} denote camera, depth and LiDAR. \textit{Vol. Size} is the volumetric size. $^1$Note that nuScenes-Occupancy has 34K key frames, where 6 images are in each frame (\ie, 200K image frames).}
\label{tab:datasets}
\vspace{-0.2cm}
\end{table*}

\vspace{-1em}
%%%%%%%%% BODY TEXT
\section{Introduction}
Accurately perceiving 3D structures of different objects and regions in urban scenes is a fundamental requirement for safe driving, thus there are growing interests in semantic occupancy perception \cite{semantickitti,nyu,nyucad,suncg,sun3d,scenenn,scannet}. Unlike 3D detection \cite{kitti,argo,lyft,nusc,waymo} and LiDAR segmentation \cite{semantickitti,waymo,panoptic} that are designed for foreground objects or sparse scanned points, the occupancy task targets at assigning semantic labels to every spatially-occupied region within the perceptive range. Therefore, semantic occupancy perception is a promising and challenging research direction in autonomous-driving perception.

Despite growing interests in semantic occupancy perception, most of the relevant benchmarks \cite{nyu,nyucad,suncg,sun3d,scenenn,scannet} are devised for indoor scenes. SemanticKITTI \cite{semantickitti} extends the occupancy perception to driving scenarios, but its dataset is relatively small in scale and limited in diversity, which hinders the generalization and evaluation of the developed occupancy perception algorithms. Besides, SemanticKITTI only evaluates the front-view occupancy predictions, while the surrounding perception is more critical for safe driving.
To address these problems, we propose OpenOccupancy, which is the first surrounding semantic occupancy perception benchmark. In the OpenOccupancy benchmark, we introduce nuScenes-Occupancy, which extends the large-scale nuScenes \cite{nusc} dataset with dense semantic occupancy annotation. As shown in Tab.~\ref{tab:datasets}, the number of annotated scenes and frames (of nuScenes-Occupancy) are $\sim$40$\times$ and $\sim$20$\times$ more than that of \cite{semantickitti}. Notably, it is almost impractical to directly annotate large-scale occupancy labels by human labor. Therefore, the \textbf{A}ugmenting \textbf{A}nd \textbf{P}urifying (AAP) pipeline is introduced to efficiently annotate and densify the occupancy labels. Specifically, we initialize annotation by multi-frame LiDAR points superimposition, where the per-point semantic labels are from \cite{panoptic}.
Considering the sparsity of the initial annotation (\ie, some occupancy labels are missed due to occlusion or limited LiDAR channels), we augment it with pseudo occupancy labels, which are constructed by the pretrained baseline (see Sec.~\ref{sec:baseline}). To further reduce noise and  artifacts, human endeavors are leveraged to purify the augmented annotation. Based on the AAP pipeline, we generate $\sim$2$\times$ dense occupancy labels than the initial annotation. Visualizations of the dense annotation are shown in Fig.~\ref{fig:showgt}.
\looseness=-1

To facilitate future research, we establish camera-based, LiDAR-based and multi-modal baselines for the OpenOccupancy benchmark. Experiment results show that the camera-based method achieves better performance on small objects (\eg, \textit{bicycle, pedestrian, motorcycle}), while the LiDAR-based approach shows superior performance on large structured regions (\eg, \textit{drivable surface, sidewalk}). Notably, the multi-modal baseline adaptively fuses intermediate features from both modalities, relatively improving the overall performance (of camera-based and LiDAR-based methods) by 47\% and 29\%. Considering the computational burden of the surrounding occupancy perception, the proposed baselines can only generate low-resolution predictions. Towards an efficient occupancy perception, we propose the Cascade Occupancy Network (CONet) that builds a coarse-to-fine pipeline upon the proposed baseline, relatively improving the performance by $\sim$30\%.
\looseness=-1

The main contributions are summarized as follows: 
    (1) We propose OpenOccupancy, which is the first benchmark designed for surrounding occupancy perception in driving scenarios.
    (2) The AAP pipeline is proposed to efficiently annotate and densify semantic occupancy labels of the nuScenes dataset, and the resulted nuScenes-Occupancy is the first dataset for surrounding semantic occupancy segmentation. 
    (3) We establish camera-based, LiDAR-based and multi-modal baselines in the OpenOccupancy benchmark. Besides, the CONet is introduced to alleviate the computational burden of high-resolution occupancy predictions, which relatively improves the baseline by $\sim$30\%.
    (4) Based on the OpenOccupancy benchmark, we conduct comprehensive experiments on the proposed baselines, CONet, and modern occupancy perception approaches.
\looseness=-1

\vspace{-0.5em}
\section{Related Work}
\vspace{-0.5em}

\noindent
\textbf{Semantic occupancy perception benchmarks.} Semantic occupancy perception originates from SUNCG \cite{suncg}, where the algorithms are required to output occupancy and semantic labels for all voxels in the camera-view frustum. In recent years, semantic occupancy perception draws growing attention and is thoroughly reviewed in \cite{sscsurvey}. To facilitate the development of occupancy perception, various relevant benchmarks have been released \cite{semantickitti,nyu,nyucad,suncg,sun3d,scenenn,scannet,semanticposs,synthcity}. Among these  benchmarks, SUNCG \cite{suncg}, NYUv2 \cite{nyu}, NYUCAD \cite{nyucad}, SUN3D \cite{sun3d}, SceneNN \cite{scenenn}, ScanNet \cite{scannet} focus on the indoor stationary scenarios. Unlike the prevalence of indoor datasets, few benchmarks \cite{synthcity,semantickitti,semanticposs,panoptic} are devised for outdoor scenes. SynthCity \cite{synthcity}, SemanticPOSS \cite{semanticposs}, Panoptic nuScenes \cite{nusc} only provide semantic labels for sparse/synthetic point clouds. SemanticKITTI \cite{semantickitti} is most relevant to the proposed OpenOccupancy benchmark, as it annotates real-world occupancy in driving scenarios. However, SemanticKITTI lacks diversity in urban scenes, which hinders the generalization of occupancy perception algorithms. Besides, it only evaluates front-view occupancy predictions.
\looseness=-1

\noindent
\textbf{Semantic occupancy perception approaches.}
Most existing occupancy perception methods rely on geometric inputs, including occupancy grids \cite{js3cnet,lmscnet,TS3D,scfusion}, LiDAR points \cite{LDIF,SPC}, RGBD images \cite{aicnet,ddrn,pial,amfn,see}, and Truncated Signed Distance Function (TSDF) \cite{loop,sketch,single,edgenet,forknet,efficient,cascaded}. MonoScene \cite{monoscene} is the first camera-based occupancy perception method in the literature, which can deduce occupancy semantics from a single image. Despite the significant development of occupancy perception approaches, most of them focus on front-view indoor scenarios. Recently, TPVFormer \cite{tpv} proposes a \textit{tri-perspective view} representation to generate surrounding occupancy prediction, yet its occupancy output is relatively sparse, as TPVFormer is designed for LiDAR segmentation.

\begin{algorithm}[tb]
  \caption{Augmenting And Purifying (AAP)}
  \label{alg:aap}
  \textbf{Input}:

  \quad $P=\{P_i\}_{i=1}^{N}\in\mathbb{R}^{M\times 3}$ are multi-frame LiDAR points.
  
  \quad $T=\{T_i\}_{i=1}^{N}\in\mathbb{R}^{N\times 3 \times 3}$ are extrinsic parameters.

  \quad $B=\{B_i\}_{i=1}^{N}$ are bounding boxes in each frame.

  \quad $S=\{S_i\}_{i=1}^{N}\in\mathbb{R}^{M}$ are semantic labels of $P$.
  
  \quad $I=\{I_i\}_{i=1}^{N}\in\mathbb{R}^{N\times 6 \times H_i \times W_i \times 3}$ are multi-frame images.

  % \quad $\mathcal{F}_m$ is the multimodal baseline (Sec.~\ref{sec:baseline}).

  \textbf{Output}: 
  
  \quad  Multi-frame occupancy ground truth $V_{\rm{final}}=\{V_i\}_{i=1}^{N}$.
  
  \begin{algorithmic}[1] %[1] enables line numbers
    \STATE{$V_{\rm{init}}=\mathcal{F}_{\rm{vox}}(\mathcal{F}_{\rm{sup}}(P,L,T,B))$} \hfill $V_{\rm{init}}\in\mathbb{R}^{N\times D\times H \times W}$
    \STATE{$\mathcal{F}_{\rm{m}}=\rm{TRAIN}$$(\mathcal{F}_{\rm{m}}(P,I), V_{\rm{init}})$}
    \STATE{$V_{\rm{pseudo}}=\mathcal{F}_{\rm{m}}(P,I)$} \hfill $V_{\rm{pseudo}}\in\mathbb{R}^{N\times D\times H \times W}$
    \STATE{$V_{\rm{aug}}=\mathcal{F}_{\rm{aug}}(V_{\rm{pseudo}},V_{\rm{init}})$} \hfill $V_{\rm{aug}}\in\mathbb{R}^{N\times D\times H \times W}$
    \STATE{$V_{\rm{final}}=\mathcal{F}_{\rm{purify}}(V_{\rm{aug}})$} \hfill $V_{\rm{final}}\in\mathbb{R}^{N\times D\times H \times W}$
    
  \end{algorithmic}
  \end{algorithm}

\section{The OpenOccupancy Benchmark}
In this section, the concept of surrounding semantic occupancy perception is first introduced.
Then we introduce nuScenes-Occupancy, which extends the nuScenes dataset \cite{nusc} with dense semantic occupancy annotations based on the AAP pipeline. Subsequently, the evaluation protocol is presented to comprehensively assess the surrounding occupancy perception algorithms.
Finally, we propose camera-based, LiDAR-based and multi-modal baselines for the OpenOccupancy Benchmark.
\subsection{Surrounding Semantic Occupancy Perception}
Referring to \cite{suncg}, surrounding semantic occupancy perception is a task for generating a complete 3D representation of volumetric occupancy and semantic labels for a scene. Different from the monocular paradigm \cite{suncg} that focuses on the front-view perception, the surrounding occupancy perception algorithms target at producing semantic occupancy in the surround-view driving scenarios. Specifically, given 360-degree inputs $X_i$ (\eg, LiDAR sweeps or surround-view images), the perception algorithms are required to predict the surrounding occupancy labels $\mathcal{F}(X_i)\in\mathbb{R}^{D\times H\times W}$, where $D,H,W$ is the volumetric size of the entire scene. It is noted that the surround-view inputs cover $\sim$5$\times$ perceptive range more than that of front-view sensors. Therefore, the core challenge of the surrounding occupancy perception lies in efficiently constructing high-resolution occupancy.

 \begin{figure}[t]
\centering
\resizebox{0.9\linewidth}{!}{
\includegraphics{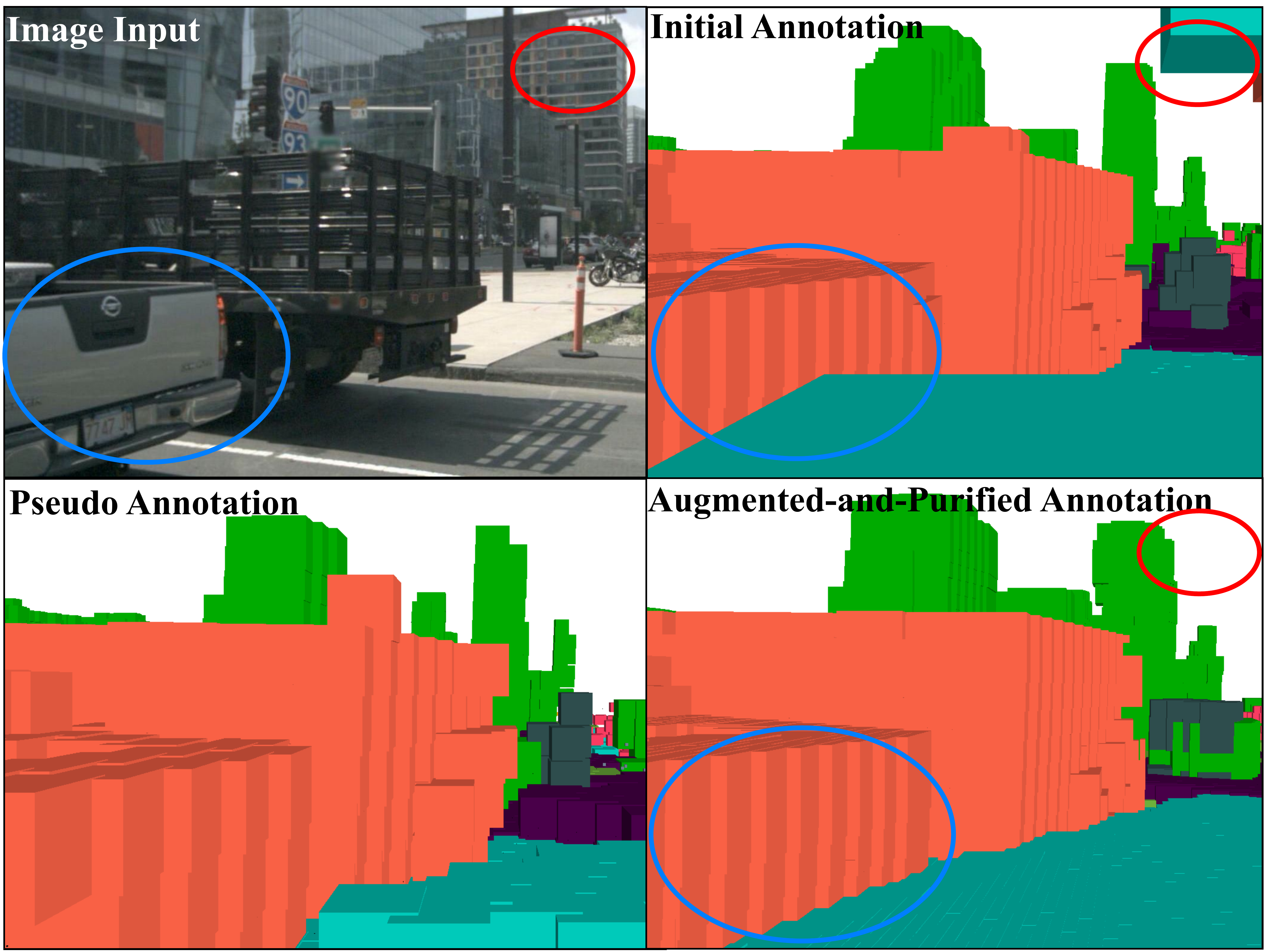}}
\caption{Comparison between the initial, pseudo and the augmented-and-purified annotation, where regions highlighted by \textcolor{red}{red} and \textcolor{blue}{blue} circle indicate that the augmented annotation is more dense and accurate.}
\label{fig:label_cmp}
\vspace{-1em}
\end{figure}

\subsection{nuScenes-Occupancy}
 SemanticKITTI \cite{semantickitti} is the first dataset for outdoor occupancy perception, but it lacks diversity in driving scenes and only evaluates front-view predictions. Towards a large-scale surrounding occupancy perception dataset, we introduce the nuScenes-Occupancy that extends the nuScenes \cite{nusc} dataset with dense semantic occupancy annotation. Although sparse LiDAR semantic labels are provided in \cite{panoptic}, it is almost unfeasible to directly annotate dense occupancy labels through human effort. Therefore, the AAP pipeline is introduced to efficiently annotate and densify the occupancy labels. 
 \looseness=-1

\begin{figure*}[ht]
\centering
\resizebox{1\linewidth}{!}{
\includegraphics{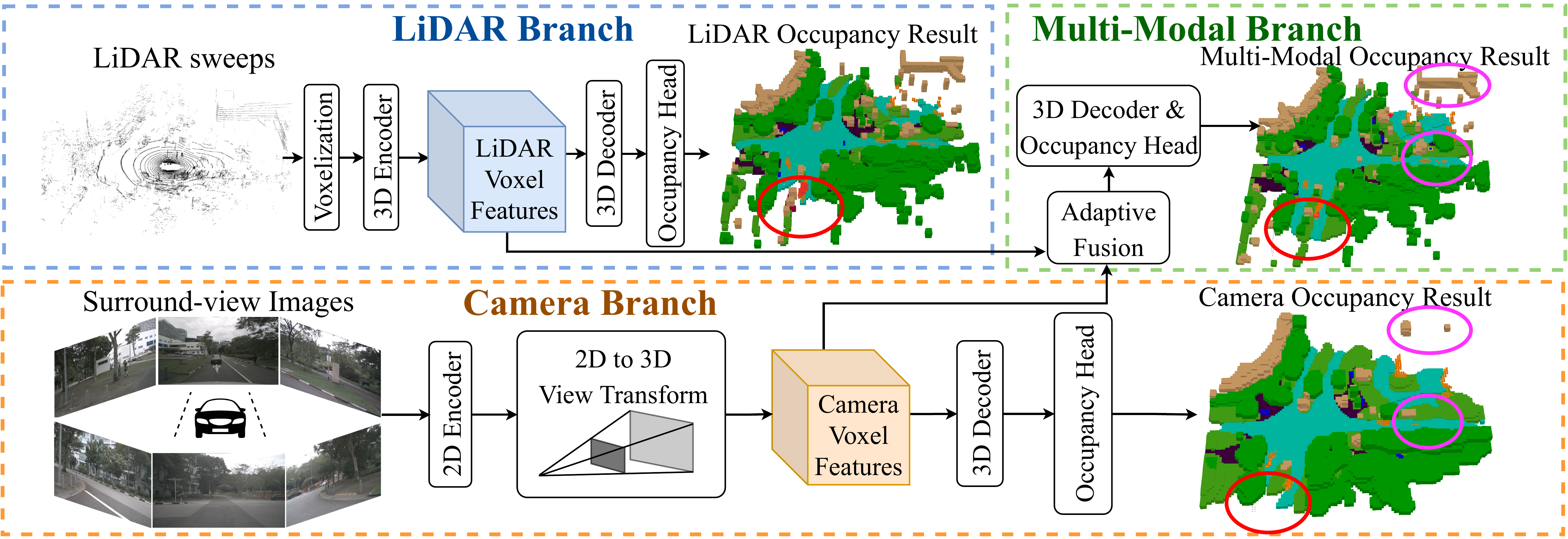}}
\caption{Overall architecture of three proposed baselines. The LiDAR branch utilizes 3D encoder to extract voxelized LiDAR features, and the camera branch uses 2D encoder to learn surround-view features, which are then transformed to generate 3D camera voxel features. In the multi-modal branch, the adaptive fusion module dynamically integrates features from two modalities. All three branches leverage 3D decoder and occupancy head to produce semantic occupancy. In the occupancy results figures, regions highlighted by \textcolor{red}{red} and \textcolor{purple}{purple} circles indicate that the multi-modal branch can generate more complete and accurate predictions (better viewed when zoomed in).}
\label{fig:mm}
\vspace{-0.5em}
\end{figure*}

 The overall AAP pipeline is shown in Alg.~\ref{alg:aap}. We first initialize annotation by LiDAR points superimposition $V_{\rm{init}}=\mathcal{F}_{\rm{vox}}(\mathcal{F}_{\rm{sup}}(P,L,T,B))$ \cite{semantickitti}, where static points (\eg, \textit{sidewalk}) are transformed to the unified world coordinate using extrinsics $T$. For movable objects (\eg, the moving \textit{car}), we transform point clouds to coordinates of their bounding boxes $B$ (each object in different frames can be associated via the instance token \cite{nudev}). Subsequently, the static and dynamic points are concatenated and voxelized ($\mathcal{F}_{\rm{vox}}$) to produce the initial occupancy annotation $V_{\rm{init}}$, where the semantic labels $S$ are form \cite{panoptic}. Note that some occupancy labels are missed due to occlusion or sparse LiDAR channels. Inspired by self-training \cite{SelfTraining}, we complement the initial annotation with pseudo occupancy labels.  Specifically, the initial annotation is utilized to train the proposed multi-modal baseline $\mathcal{F}_{\rm{m}}$ (see Sec.~\ref{sec:baseline}), and pseudo occupancy labels $V_{\rm{pseudo}}$ are produced by the pretrained model. Then we augment initial labels with pseudo labels to construct dense annotations $V_{\rm{aug}}=\mathcal{F}_{\rm{aug}}(V_{\rm{pseudo}},V_{\rm{init}})$. To resolve conflicts in the two annotations, we only augment empty voxels in $V_{\rm{init}}$:
 \begin{small}
     \begin{equation}
     V_{\rm{aug}}(x,y,z) = \left\{\begin{array}{ll}
            V_{\rm{init}}(x,y,z)  & V_{\rm{init}}(x,y,z) \text{ is occupied}\\
            V_{\rm{pseudo}}(x,y,z) & \text{else}.
         \end{array}\right.
 \end{equation} 
 \end{small}

 Regarding artifacts caused by pseudo labels, human endeavors are further leveraged to purify the augmented labels and establish final annotation $V_{\rm{final}}=\mathcal{F}_{\rm{purify}}(V_{\rm{aug}})$. For efficiency, labeling software is devised for human annotators, where the 3D semantic occupancy is projected to multi-view images, and annotators can efficiently determine the occupancy boundary through both 3D global view and 2D camera views (the purifying process involves $\sim$4000 human hours of labeling effort). 
 \looseness=-1
 
 As shown in Fig.~\ref{fig:label_cmp}, the pseudo labels are complementary to the initial annotation, and the augmented-and-purified labels are more dense and precise.  Notably, $\sim$400K occupied voxels are in each frame of the augmented-and-purified annotation, which is $\sim$2$\times$ dense than the initial annotation.  In summary, nuScenes-Occupancy has 28130 training frames and 6019 validation frames, where 17 semantic labels (same as \cite{panoptic}) are assigned to occupied voxels in each frame.

\subsection{Evaluation Protocol}
The evaluation range is set as $[-51.2\rm{m},51.2\rm{m}]$ for $X,Y$ axis, and $[-3\rm{m},5\rm{m}]$ for $Z$ axis. Following \cite{semantickitti}, the voxel resolution is $0.2\rm{m}$, which results in a volume of $40\times 512\times 512$ voxels for occupancy prediction. For evaluation metrics, we utilize Intersection of Union (IoU) \cite{semantickitti} as the \textit{geometric metric}, which identifies a voxel as being occupied or empty (\ie, deem all occupied voxels as one category):
\begin{equation}
    \rm{IoU} =  \frac{\mathrm{TP}_{o}}{\mathrm{TP}_{o}+\mathrm{FP}_{o}+\mathrm{FN}_{o}},
\end{equation}
where $\mathrm{TP}_{o},\mathrm{FP}_{o},\mathrm{FN}_{o}$ are the number of true positive, false positive and false negative predictions for occupied voxels.
Besides, we calculate the mean IoU (mIoU) of each class as the \textit{semantic metric}:
\begin{equation}
    \rm{mIoU} = \frac{1}{C_{\rm{sem}}} \sum_{c=1}^{C_{\rm{sem}}} \frac{\mathrm{TP}_{c}}{\mathrm{TP}_{c}+\mathrm{FP}_{c}+\mathrm{FN}_{c}},
\end{equation}
where $\mathrm{TP}_{c},\mathrm{FP}_{c},\mathrm{FN}_{c}$ denote the number of true positive, false positive and false negative predictions for class $c$, and $C_{\rm{sem}}$ is the total number of classes. Following \cite{panoptic}, the \textit{noise} class \cite{panoptic} is ignored in the evaluation.

\begin{figure*}[ht]
\centering
\resizebox{1\linewidth}{!}{
\includegraphics{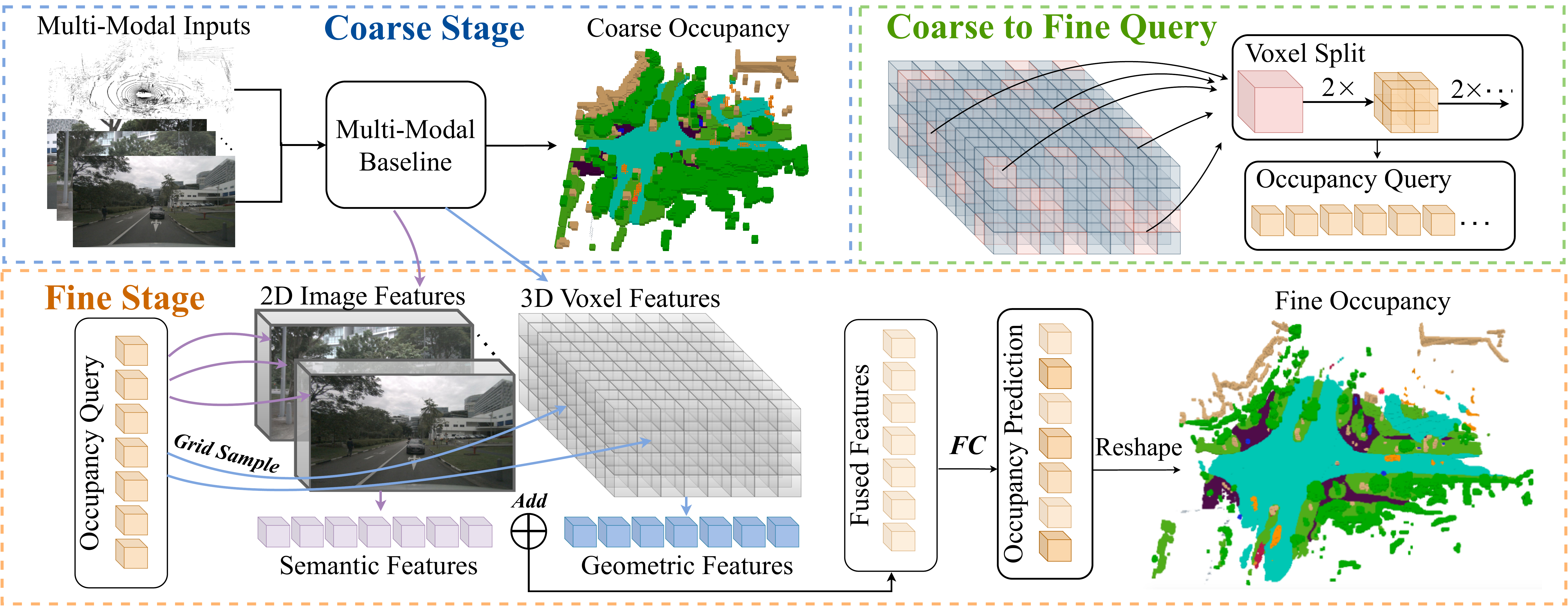}}
\caption{Overall framework of the multi-modal CONet. (1) The coarse occupancy is first generated by the multi-modal baseline. (2) Then the occupied voxels are split to produce high-resolution occupancy queries. (3) Subsequently, we project queries to sample from 2D image features and 3D voxel features. The sampled features are fused and regularized by Fully-Connected (FC) layers to generate fine-grained occupancy predictions.}
\label{fig:conet}
\vspace{-1em}
\end{figure*}

\subsection{OpenOccupancy Baselines}
\label{sec:baseline}
The majority of existing occupancy perception methods \cite{aicnet,ddrn,pial,amfn,see,suncg,loop,sketch,single,edgenet,forknet,monoscene} are designed for front-view perception.
To extend these approaches to surrounding occupancy perception, each camera-view input is processed individually, which is inefficient. Besides,  inconsistency may exist in the overlap region of two adjacent outputs. To mitigate these problems, we establish baselines that coherently learn surrounding semantic
occupancy from 360-degree inputs (\eg, LiDAR sweeps or surround-view images). Specifically, camera-based, LiDAR-based and multi-modal baselines are proposed for the OpenOccupancy benchmark.

\noindent
\textbf{LiDAR-based baseline.} As shown in the top-left diagram of Fig.~\ref{fig:mm}, parameterized voxelization \cite{zhou2018voxelnet} is first utilized to embed raw LiDAR points to voxelized features.
% $F^{\mathcal{L}_p}\in\mathbb{R}^{X\times Y \times Z \times C^\mathcal{L}}$ ($X, Y, Z$ are volumetric sizes, and $C^\mathcal{L}$ is the feature channel)
For computational efficiency, 3D sparse convolutions \cite{SECONDSE} are leveraged to encode features in the voxel space, producing LiDAR voxel features $F^{\mathcal{L}}$ with reduced spatial dimension ($\frac{D}{S}\times \frac{H}{S} \times \frac{W}{S}$, $S$ is the stride). The voxel features are further decoded by 3D convolutions, generating multi-scale voxel features $F^\mathcal{L}_i\in\mathbb{R}^{\frac{D}{ 2^iS}\times \frac{H}{2^iS} \times \frac{W}{2^iS} \times C_i} (i=0,1,2)$. These features are upsampled and concatenated along the channel dimension, resulting in $F^\mathcal{L}_{\rm{ms}}\in\mathbb{R}^{\frac{D}{S}\times \frac{H}{S} \times \frac{W}{S} \times \sum_{i=0}^2C_i}$. Finally, the occupancy head is utilized to reduce feature channels, and a \textit{softmax} function is leveraged to produce semantic probabilities. The output $O^\mathcal{L}\in\mathbb{R}^{\frac{D}{S}\times \frac{H}{S} \times \frac{W}{S} \times 18}$ (18: 1 empty label with 17 semantic labels in nuScenes-Occupancy) can be scaled to arbitrary sizes using the \textit{trilinear interpolation}, and class labels can be determined by the \textit{argmax} function along the channel dimension.
\looseness=-1

\noindent
\textbf{Camera-based baseline.} As illustrated in the bottom of Fig.~\ref{fig:mm}, the 2D encoder (\eg, ResNet \cite{resnet} and FPN \cite{fpn}) is first utilized to extract multi-view features $F^{mv}$. Subsequently, we apply the \textit{2D to 3D view transform} \cite{bevfusion} to project 2D features into 3D ego-car coordinates. Different from \cite{bevfusion} that collapses 3D features onto the Bird's Eye View (BEV) plane, the height information is reserved for a fine-grained 3D occupancy prediction. The resulted camera voxel features $F^\mathcal{C}$ have the same volumetric size as that of $F^\mathcal{L}$. Following the LiDAR-based baseline, we further employ the 3D decoder and occupancy head to output the semantic occupancy $O^\mathcal{C}\in\mathbb{R}^{\frac{D}{S}\times \frac{H}{S} \times \frac{W}{S} \times 18}$.
\looseness=-1

\noindent
\textbf{Multi-modal baseline.} The LiDAR voxel features $F^\mathcal{L}$ and camera voxel features $F^\mathcal{C}$ are natural representations for occupancy prediction. In the multi-modal baseline, we propose the adaptive fusion module to dynamically integrate features from $F^\mathcal{L}$ and $F^\mathcal{C}$:
\begin{align}
        W &= \mathcal{G}_{\rm{C}}(\left[\mathcal{G}_{\rm{C}}(F^\mathcal{L}), \mathcal{G}_{\rm{C}}(F^\mathcal{C})\right]),\\
        F^\mathcal{F}&=\sigma(W)\odot F^\mathcal{L}+(1-\sigma(W))\odot F^\mathcal{C},
\end{align}
where $\mathcal{G}_{\rm{C}}$ is the 3D convolution, $[\cdot,\cdot]$ is the concatenation along feature channel, $\sigma$ denotes \textit{Sigmoid} function and $\odot$ represents element-wise product. Based on the fused voxel features $F^\mathcal{F}$, the final occupancy can be predicted by the aforementioned 3D decoder and occupancy head. 

To train the proposed baselines, cross-entropy loss $\mathcal{L}_{\rm{ce}}$ and lovasz-softmax loss $\mathcal{L}_{\rm{ls}}$ \cite{lovasz} are leveraged to optimize the network. Following \cite{monoscene}, we also utilize affinity loss $\mathcal{L}_{\rm{scal}}^{\rm{geo}}$ and $\mathcal{L}_{\rm{scal}}^{\rm{sem}}$ to optimize the scene-wise and class-wise metrics (\ie, geometric IoU and semantic mIoU). Besides, the explicit depth supervision $\mathcal{L}_{\rm{d}}$ \cite{bevdepth} is used to train a depth-aware \textit{view transform} module. Therefore, the overall loss function can be derived as:
\begin{equation}
    \mathcal{L}_{\rm{total}} = \mathcal{L}_{\rm{ce}} + \mathcal{L}_{\rm{ls}} + \mathcal{L}_{\rm{scal}}^{\rm{geo}} + \mathcal{L}_{\rm{scal}}^{\rm{sem}} + \mathcal{L}_{\rm{d}},
\end{equation}
where $\mathcal{L}_{\rm{d}}$ is only calculated in the camera-based and multi-modal baseline.

\begin{table*}
	\setlength{\tabcolsep}{0.0035\linewidth}
	\newcommand{\classfreq}[1]{{~\tiny(\semkitfreq{#1}\%)}}  %
	\centering
   \resizebox{1\linewidth}{!}{
	\begin{tabular}{l|c c | c c | c c c c c c c c c c c c c c c c}
 
		\toprule
		Method
		& \makecell[c]{Input}
		& \makecell[c]{Surround}
		& \makecell[c]{IoU}
            & \makecell[c]{mIoU}
		& \rotatebox{90}{\textcolor{barrier}{$\blacksquare$} barrier} 
		& \rotatebox{90}{\textcolor{bicycle}{$\blacksquare$} bicycle}
		& \rotatebox{90}{\textcolor{bus}{$\blacksquare$} bus} 
		& \rotatebox{90}{\textcolor{car}{$\blacksquare$} car} 
		& \rotatebox{90}{\textcolor{const. veh.}{$\blacksquare$} const. veh.} 
		& \rotatebox{90}{\textcolor{motorcycle}{$\blacksquare$} motorcycle} 
		& \rotatebox{90}{\textcolor{pedestrian}{$\blacksquare$} pedestrian} 
		& \rotatebox{90}{\textcolor{traffic cone}{$\blacksquare$} traffic cone} 
		& \rotatebox{90}{\textcolor{trailer}{$\blacksquare$} trailer} 
		& \rotatebox{90}{\textcolor{truck}{$\blacksquare$} truck} 
		& \rotatebox{90}{\textcolor{drive. suf.}{$\blacksquare$} drive. suf.} 
		& \rotatebox{90}{\textcolor{other flat}{$\blacksquare$} other flat} 
		& \rotatebox{90}{\textcolor{sidewalk}{$\blacksquare$} sidewalk} 
		& \rotatebox{90}{\textcolor{terrain}{$\blacksquare$} terrain} 
		& \rotatebox{90}{\textcolor{manmade}{$\blacksquare$} manmade} 
		& \rotatebox{90}{\textcolor{vegetation}{$\blacksquare$} vegetation} \\
		% & mIoU\\
		\midrule
		MonoScene~\cite{monoscene} & C & \xmark  & 18.4 & 6.9 & 7.1  & 3.9  &  9.3 &  7.2 & 5.6  & 3.0  &  5.9& 4.4& 4.9 & 4.2 & 14.9 & 6.3  & 7.9 & 7.4  & 10.0 & 7.6 \\
  
  		TPVFormer~\cite{tpv} &C &  \cmark& 15.3 &  7.8 & 9.3  & 4.1  &  11.3 &  10.1 & 5.2  & 4.3  & 5.9 & 5.3&  6.8& 6.5 & 13.6 & 9.0  & 8.3 & 8.0  & 9.2 & 8.2 \\
    
            3DSketch~\cite{sketch} &  C\&D & \xmark& 25.6 & 10.7  & 12.0 &  5.1 &  10.7 &  12.4 & 6.5  & 4.0  & 5.0 & 6.3&  8.0&  7.2& 21.8 &  14.8 & 13.0 &  11.8 & 12.0 & 21.2 \\
            
            AICNet~\cite{aicnet} & C\&D   &  \xmark& 23.8 & 10.6  & 11.5  & 4.0  & 11.8  & 12.3&  5.1 & 3.8  & 6.2  & 6.0 & 8.2&  7.5&  24.1 & 13.0 & 12.8  & 11.5 & 11.6  &  20.2\\

            LMSCNet~\cite{lmscnet} & L &  \cmark& 27.3 & 11.5 & 12.4&  4.2 & 12.8  & 12.1  & 6.2  &  4.7 & 6.2 & 6.3&  8.8&  7.2& 24.2 & 12.3  & 16.6 & 14.1  & 13.9 & 22.2 \\

		JS3C-Net~\cite{js3cnet} &L &  \cmark& 30.2  & 12.5 & 14.2 & 3.4  & 13.6  & 12.0  & 7.2  &  4.3 & 7.3 & 6.8&  9.2& 9.1 & 27.9 & 15.3  & 14.9 & 16.2  & 14.0 & \textbf{24.9} \\
		\midrule
            C-baseline (ours)  & C &  \cmark&19.3  & 10.3  &  9.9 & 6.8  & 11.2  & 11.5  & 6.3  & 8.4  & 8.6 & 4.3 & 4.2 & 9.9 & 22.0  & 15.8 & 14.1  & 13.5  & 7.3&10.2 \\
            
            L-baseline (ours)& L &  \cmark& 30.8  & 11.7 &  12.2  & 4.2  & 11.0  & 12.2  & 8.3  & 4.4  & 8.7 & 4.0& 8.4 & 10.3 & 23.5& 16.0 & 14.9 & 15.7  & 15.0 &17.9  \\

            M-baseline (ours)& C\&L &   \cmark& 29.1 & 15.1 &  14.3  & 12.0  & 15.2  & 14.9  & 13.7  & 15.0  & 13.1 & 9.0 & 10.0 & 14.5 & 23.2 & 17.5 & 16.1  & 17.2 & 15.3  & 19.5  \\
		\midrule
            C-CONet (ours) & C &  \cmark&20.1  & 12.8&13.2  & 8.1 &  15.4 &  17.2 & 6.3  & 11.2  & 10.0  &  8.3 & 4.7 & 12.1 & 31.4 & 18.8 & 18.7  & 16.3 & 4.8  &8.2  \\
            
            L-CONet (ours) & L &  \cmark& \textbf{30.9}  & 15.8 &  17.5  & 5.2  & 13.3  & 18.1  & 7.8  & 5.4  & 9.6 & 5.6& 13.2 & 13.6 & \textbf{34.9} & \textbf{21.5}  & 22.4 & \textbf{21.7}  & 19.2 &23.5  \\

            M-CONet (ours) & C\&L &   \cmark& 29.5 & \textbf{20.1} &  \textbf{23.3}  & \textbf{13.3}  & \textbf{21.2}  & \textbf{24.3}  & \textbf{15.3}  & \textbf{15.9}  & \textbf{18.0} & \textbf{13.3} & \textbf{15.3} &\textbf{ 20.7} & 33.2 & 21.0 & \textbf{22.5}  & 21.5 & \textbf{19.6}  & 23.2  \\
		\bottomrule
	\end{tabular}}\\

	\caption{Performance on nuScenes-Occupancy (validation set). We report the geometric metric IoU, semantic metric mIoU, and the IoU for each semantic class. The $C,D,L,M$ denotes \textit{camera, depth, LiDAR} and \textit{multi-modal}. For \textit{Surround=}\cmark, the method directly predicts surrounding semantic occupancy with 360-degree inputs. Otherwise, the method produces the results of each camera view, and then concatenates them as surrounding outputs. }
	\label{table:base_main}
\end{table*}

\section{Cascade Occupancy Network}

Compared with front-view occupancy perception \cite{semantickitti}, the input of the surrounding occupancy perception covers $\sim$5$\times$ perceptive range. Therefore, the complexity lies in the computational burden of high-resolution 3D prediction. For efficiency, the stride parameter $S$ is set as 4 in the proposed baselines (\ie, the volumetric size of the output is $(10\times 128\times 128)$). Notably, we empirically find that using a smaller stride parameter (\eg, S=2) enhances the performance. However, the GPU memory is approximately 2$\times$ upscaled ($\sim$40~GB in the training phase). Therefore, we propose the Cascade Occupancy Network for an efficient yet accurate surrounding occupancy perception.

Specifically, CONet introduces a coarse-to-fine pipeline, which can be efficiently built upon the proposed baselines. Taking the multi-modal baseline for example  (termed as multi-modal CONet), the  overall framework is shown in Fig.~\ref{fig:conet}. The coarse occupancy $O^\mathcal{M}\in\mathbb{R}^{\frac{D}{S}\times \frac{H}{S} \times \frac{W}{S} \times 18}$ is first generated by the multi-modal baseline, where the occupied voxels $V_{\rm{o}}\in\mathbb{R}^{N_{\rm{o}}\times 3}$ ($N_{\rm{o}}$ is the number of occupied voxels, and 3 denotes the $(x,y,z)$ indices in voxel coordinates) are split as high-resolution occupancy queries $Q_{\rm{H}}\in\mathbb{R}^{N_{\rm{o}}8^{\eta-1}\times 3}$:
\begin{equation}
    Q_{\rm{H}} = \mathcal{T}_{\rm{v}\rightarrow \rm{w}}(\mathcal{F}_{\rm{s}}(V_{\rm{o}}, \eta)),
\end{equation}
where $\mathcal{F}_{\rm{s}}$ is the voxel split function (\ie, for $(x_0,y_0,z_0)$ in $V_{\rm{o}}$, the split indices are $\{x_0+\frac{i}{\eta},y_0+\frac{j}{\eta},z_0+\frac{k}{\eta}\}(i,j,k\in(0,\eta-1))$), $\eta$ is the split ratio (typically set as 4), and $\mathcal{T}_{\rm{v}\rightarrow \rm{w}}$ transforms the voxel coordinates to the world coordinates. Subsequently, we project $Q_{\rm{H}}$ on 2D image plane to sample semantic features $F^{\mathcal{S}}=\mathcal{G}_{\rm{S}}(F^{mv},\mathcal{T}_{\rm{w}\rightarrow\rm{c}}(Q_{\rm{H}}))$, and transform $Q_{\rm{H}}$ to voxel space to sample geometric features $F^{\mathcal{G}}=\mathcal{G}_{\rm{S}}(F^{\mathcal{F}},\mathcal{T}_{\rm{w}\rightarrow\rm{v}}(Q_{\rm{H}}))$ ($\mathcal{G}_{\rm{S}}$ is the \textit{grid sample} function \cite{stn}, $\mathcal{T}_{\rm{w}\rightarrow\rm{c}}$ and $\mathcal{T}_{\rm{w}\rightarrow\rm{v}}$ are transformations from world coordinates to camera coordinates and voxel coordinates). The sampled features are then fused and regularized by FC layers to produce fine-grained occupancy predictions:
\begin{equation} 
    O^{\rm{fg}} = \mathcal{G}_f(\mathcal{G}_f(F^{\mathcal{S}})+\mathcal{G}_f(F^{\mathcal{G}})),
\end{equation}
where $F^{\mathcal{G}}$ are FC layers. Finally, $O^{\rm{fg}}$ can be reshaped to the volumetric representation $O^{\rm{vol}}\in\mathbb{R}^{\frac{\eta D}{S}\times \frac{\eta H}{S} \times \frac{\eta W}{S} \times 18}$:
\begin{footnotesize}
\begin{equation}
  O^{\rm{vol}}(x,y,z) = \left\{\begin{array}{ll}
    O^{\rm{fg}}(\mathcal{T}_{\rm{v}\rightarrow\rm{q}}(x,y,z)) & (x,y,z)\in \mathcal{T}_{\rm{w}\rightarrow\rm{v}}(Q_{\rm{H}})\\
    \text{Empty~Label} & (x,y,z)\notin \mathcal{T}_{\rm{w}\rightarrow\rm{v}}(Q_{\rm{H}}),
    \end{array}\right.
\end{equation}
\end{footnotesize}
where $\mathcal{T}_{\rm{v}\rightarrow\rm{q}}$ transforms the voxel coordinates to indices of the high-resolution query $Q_{\rm{H}}$. Notably, the CONet can also be generalized to camera-based and LiDAR-based baselines. For camera-based CONet, we sample $Q_{\rm{H}}$ from $F^{mv}$ and $F^{\mathcal{C}}$. For LiDAR-based CONet that without multi-view 2D features, we only sample $Q_{\rm{H}}$ from $F^{\mathcal{L}}$.

For optimization, we use the sample pipeline as that of baselines, except that the training losses are calculated on both (coarse and fine) predictions.

\section{OpenOccupancy Experiment}
In this section, the experiment setup is first given. Then we delve into surrounding occupancy assessment, including camera-based methods, LiDAR-based methods and multi-modal methods. In the next step, we analyze the baseline performance under different experiment settings. Finally, the efficiency and effectiveness of CONet are investigated. 
\looseness=-1

 \begin{figure*}[ht]
\centering
\resizebox{1\linewidth}{!}{
\includegraphics{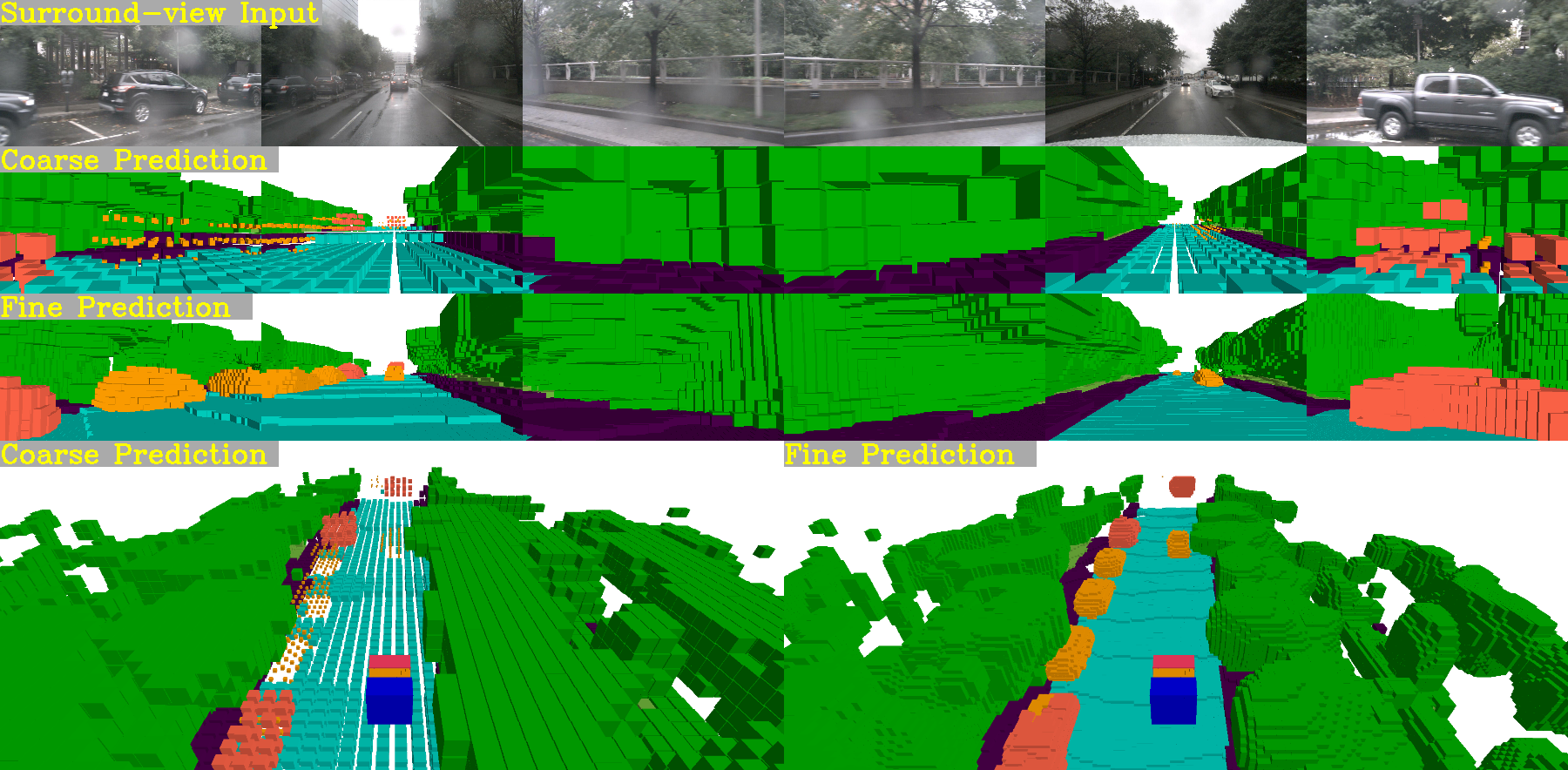}}
\caption{Visualization of the semantic occupancy predictions, where the 1\textit{st} row is surround-view images. In 2\textit{nd} and 3\textit{rd} rows, we show the camera view of coarse and fine occupancy generated by the multi-modal baseline and multi-modal CONet. In the 4\textit{th} row, we compare their global-view predictions.}
\label{fig:conet_vis}
\vspace{-0.5em}
\end{figure*}

\subsection{Experiment Setup}
In the OpenOccupancy benchmark, we evaluate surrounding semantic occupancy segmentation performance based on the nuScenes-Occupancy, where comprehensive experiments are conducted on the proposed baselines, CONet and modern occupancy perception algorithms \cite{monoscene,aicnet,tpv,sketch,js3cnet,lmscnet}. For single-view methods \cite{monoscene,sketch,aicnet}, the occupancy predictions are generated for each view individually, then we concatenate the predictions to produce surrounding occupancy results. To provide dense depth maps for \cite{sketch,aicnet}, we first project LiDAR points on the image, which is then densified by depth completion \cite{ku2018defense}. For LiDAR-based methods \cite{js3cnet,lmscnet}, we use 10 LiDAR sweeps as input. For camera-based methods, the input image size is $1600\times 900$. Unless specified, we use ImageNet \cite{imagenet} pretrained ResNet50 \cite{resnet} as the backbone for image-based baselines. Considering the output size may be smaller than that of the ground truth $(40\times 512 \times 512)$, \textit{trilinear interpolation} is utilized to upsample these outputs before evaluation. For training, we leverage the AdamW \cite{adam} optimizer with
a weight decay of 0.01 and an initial learning rate of 2e-4. We adopt the cosine learning rate scheduler with linear warming up in the first 500 iterations, and a similar augmentation strategy as BEVDet \cite{bevdet}. All models are trained for 24 epochs with a batch size of 8 on 8 A100 GPUs.
\looseness=-1

\subsection{Surrounding Occupancy Assessment}
Equipped with the OpenOccupancy benchmark, we analyze the surrounding occupancy perception performance of six modern approaches
(MonoScene \cite{monoscene}, TPVFormer \cite{tpv}, 3DSketch \cite{sketch}, AICNet \cite{aicnet}, LMSCNet \cite{lmscnet}, JS3C-Net \cite{js3cnet} and the proposed baselines and CONet.
From the results in Tab.~\ref{table:base_main}, it can be observed that:

\noindent
\textbf{(1) Compared with single-view methods, the surrounding occupancy perception paradigm shows superior performance.} Specifically, the proposed camera-based baseline and TPVFormer relatively improve MonoScene 49\% and 13\% on mIoU. Besides, the LiDAR-based baseline and surrounding occupancy perception methods \cite{lmscnet,js3cnet} surpass the RGBD paradigms \cite{aicnet,sketch} on both IoU and mIoU. Therefore, it is promising to develop surrounding occupancy perception approaches on the OpenOccupancy benchmark.

\noindent
\textbf{(2) The proposed baselines show adaptability and scalability for the surrounding occupancy perception.} For the camera-based methods, our baseline relatively improves TPVFormer by 26\% and 32\% on IoU and mIoU. For the LiDAR-based methods, our baseline outperforms LMSCNet and is comparable to JS3C-Net (Note that JS3C-Net is a two-stage method). Additionally, the proposed baselines explicitly optimize the network in a unified voxel representation, which can be naturally extended for multi-modal fusion. Consequently, the proposed multi-modal baseline relatively enhances 3DSketch, AICNet, LMSCNet, and JS3C-Net by 41\%, 42\%, 31\%, and 21\% on mIoU. 
\looseness=-1

\noindent
\textbf{(3) Information from the camera and LiDAR are complementary to each other, and the multi-modal baseline significantly enhances the performance.} Experiment results show that the LiDAR-based approach shows superior performance on large structured regions (\eg, \textit{drivable surface, sidewalk, vegetation}), while the camera-based baseline gains better performance on small objects (\eg, \textit{bicycle, pedestrian, motorcycle, traffic cone}). Notably, the multi-modal baseline adaptively fuses intermediate features from both modalities, relatively enhancing the LiDAR-based and camera-based baseline by 47\% and 29\% on mIoU.

\noindent
\textbf{(4) The complexity of surrounding occupancy perception lies in the computational burden of high-resolution 3D predictions, which can be alleviated by the proposed CONet.} The volumetric size  ($40\times 512\times 512$) of the ground truth occupancy in our benchmark is $\sim$5$\times$ larger than that of \cite{semantickitti}, and directly predicting high-resolution occupancy is computationally unfeasible. For efficiency, the proposed baselines produce low-resolution results, yet the performance is restricted. Therefore, we propose CONet to efficiently refine the low-resolution prediction. Notably, the CONet built upon camera-based, LiDAR-based and multi-modal baselines relatively improves the mIoU by 24\%, 35\% and 33\% with marginal latency overhead (efficiency comparison is in Tab.~\ref{tab:abl_conet}). Additionally, we provide visualization (see Fig.~\ref{fig:conet_vis}) to verify that the CONet can generate fine-grained occupancy results based on coarse predictions.
\looseness=-1

\begin{table}[t]
  \centering
  \resizebox{0.48\textwidth}{!}{
    \begin{tabular}{c|c|c|c|c|c}
        %%%%%%%%%%%%%%%%%%%%%%%%%%%%%%%%%%%%%%%
         \hline
        %%%%%%%%%%%%%%%%%%%%%%%%%%%%%%%%%%%%%%%%
          Method &  2D Backbone &  Input Size &  Fusion & IoU & mIoU\\
        \hline\hline
       \noalign{\smallskip}
          C & R-50 & $704\times 256$ & - &16.6 &8.6 \\ 
          C & R-50 & $1600\times 900$ & - & 19.3 & 10.3\\ 
          C & R-101 & $1600\times 900$ & - & 20.2 & 11.4\\ 
        \noalign{\smallskip}
        \hline
        \noalign{\smallskip}
          L & - & 1 sweep & - & 21.6 & 11.1\\ 
          L & - & 10 sweeps & - & 30.8 & 11.7\\ 
        \noalign{\smallskip}
        \hline
        \noalign{\smallskip}
          M &  R-50  & \makecell[c]{$1600\times 900$ \\10 sweeps} & Cat. & 28.5& 14.3\\ 
          
          M &  R-50  & \makecell[c]{$1600\times 900$ \\10 sweeps} & Add. & 28.7& 14.4\\ 
          
          M &  R-50  & \makecell[c]{$1600\times 900$ \\10 sweeps} & Adaptive & 29.1& 15.1 \\ 
          
         \noalign{\smallskip}
        %%%%%%%%%%%%%%%%%%%%%%%%%%%%%%%%%%%%%%%%
         \hline
        %%%%%%%%%%%%%%%%%%%%%%%%%%%%%%%%%%%%%%%%
    \end{tabular}}
  \caption{Ablation study on the proposed baselines, where \textit{C,L,M} denotes camera, LiDAR and multi-modal, and \textit{Cat.} represents the \textit{concatenation}.} 
    \label{tab:abl_base} % end caption
    \vspace{-1em}
\end{table}

\subsection{Baselines under Different Settings}

% \noindent
% \textbf{Ablation study on baselines.}
In this subsection, we analyze baseline performance under different experiment settings (\eg, input size, backbone selection, fusion method), and the results are shown in Tab.~\ref{tab:abl_base}. For the camera-based baseline, using a larger input size ($1600\times 900$) relatively improves IoU and mIoU by 16\% and 20\%. Besides, replacing ResNet50 with ResNet101 further enhances mIoU by 11\%. For the LiDAR-based baseline, it is observed that utilizing multi-sweeps as input (following \cite{cenerpoint,SECONDSE,pointpillar}, 10 sweeps are used) relatively improves the single-sweep counterpart by 43\% and 5\% on IoU and mIoU. For the multi-modal baseline, the \textit{concatenation} and \textit{add} operations are suboptimal for feature fusion. In contrast, the proposed adaptive fusion dynamically integrates features from two modalities, which relatively enhances the mIoU by 6\% and 5\%.
\looseness=-1

\begin{table}[t]
  \centering
  \resizebox{0.48\textwidth}{!}{
    \begin{tabular}{c|c|c|c|c}
        %%%%%%%%%%%%%%%%%%%%%%%%%%%%%%%%%%%%%%%
         \hline
        %%%%%%%%%%%%%%%%%%%%%%%%%%%%%%%%%%%%%%%%
          Method &  GPU Mem. &  GFLOPs &  IoU & mIoU\\
        \hline\hline
       \noalign{\smallskip}
          C-baseline ($S=4$) & 17~GB & 2241 & 19.3& 10.3\\ 
          C-baseline ($S=2$) &  35~GB & 6677 & 20.0& 12.2\\ 
          C-CONet & 22~GB & 2371 & 20.1&12.8 \\ 
        \noalign{\smallskip}
        \hline
        \noalign{\smallskip}
          L-baseline ($S=4$) & 7.5~GB & 749 & 30.8&11.7 \\ 
          L-baseline ($S=2$) & 22~GB & 5899 &30.7 &15.0 \\ 
          L-CONet& 8.5~GB & 810 &30.9& 15.8 \\ 
        \noalign{\smallskip}
        \hline
        \noalign{\smallskip}
          M-baseline ($S=4$) & 19~GB & 3050 & 28.9& 15.1\\ 
          M-baseline ($S=2$) & 40~GB & 13117 & 29.3&19.8 \\ 
          M-CONet& 24~GB & 3066 &29.5&20.1 \\ 
          
         \noalign{\smallskip}
        %%%%%%%%%%%%%%%%%%%%%%%%%%%%%%%%%%%%%%%%
         \hline
        %%%%%%%%%%%%%%%%%%%%%%%%%%%%%%%%%%%%%%%%
    \end{tabular}}
  \caption{Efficiency analysis on CONet, where \textit{C,L,M} denotes camera, LiDAR and multi-modal, \textit{GPU Mem.} represents the GPU memory consumption at training phase, and $S$ is the stride parameter that controls the output size.} 
    \label{tab:abl_conet} % end caption
\end{table}

\begin{table}[t]
  \centering
  \resizebox{0.45\textwidth}{!}{
    \begin{tabular}{c|c|c|c|c}
        %%%%%%%%%%%%%%%%%%%%%%%%%%%%%%%%%%%%%%%
         \hline
        %%%%%%%%%%%%%%%%%%%%%%%%%%%%%%%%%%%%%%%%
          Method &  Sem. Feat. &  Geo.Feat. &  IoU & mIoU\\
        \hline\hline
       \noalign{\smallskip}
          M-baseline & - & - & 29.1 & 15.1\\ 
          M-CONet &\cmark &  & 27.4 & 12.1\\ 
          M-CONet &  &\cmark  &29.2 & 19.3\\ 
          M-CONet & \cmark &  \cmark  &29.5 & 20.1\\ 
         \noalign{\smallskip}
        %%%%%%%%%%%%%%%%%%%%%%%%%%%%%%%%%%%%%%%%
         \hline
        %%%%%%%%%%%%%%%%%%%%%%%%%%%%%%%%%%%%%%%%
    \end{tabular}}
  \caption{Ablation study on feature sampling strategies of the CONet. \textit{M} represents multi-modal, \textit{Sem. Feat.} and \textit{Geo. Feat.} denotes semantic features and geometric features.} 
    \label{tab:abl_conet2} % end caption
    \vspace{-1.5em}
\end{table}

\subsection{Efficiency and Effectiveness of CONet} 
For efficiency, the proposed baselines generate low-resolution predictions (\ie, the stride parameter $S$ is set as 4, and the output volumetric size is $(10\times 128 \times 128)$). As shown in Tab.~\ref{tab:abl_conet}, using a smaller stride parameter (\eg, $S$=2) enhances the performance, yet the training-time GPU memory is $\sim$2$\times$ upscaled, and GFLOPs are $\sim$8$\times$ upscaled. Therefore, we propose the CONet for efficient surrounding occupancy perception. Compared with high-resolution baselines ($S$=2), the CONet built upon low-resolution baselines ($S$=4) achieves better performance on all the metrics. Besides, the CONet reduces $\sim$15~GB training-time GPU memory, and relatively decreases GFLOPs by $\sim$70\%.
Additionally, we conduct ablation study to investigate the effectiveness of the feature sampling strategy in CONet. 
 As shown in Tab.~\ref{tab:abl_conet2}, solely sampling from $F^{\mathcal{S}}$ degrades the performance, as 2D semantic features are insufficient for high-resolution 3D predictions. In contrast, sampling from geometric features $F^{\mathcal{G}}$ can improve the baseline by 28\% on mIoU. Notably, combining the two features further enhances the performance, which relatively improves the baseline by 33\%.
 \looseness=-1

\vspace{-0.3em}
\section{Conclusion}
\vspace{-0.3em}
In this paper, we propose OpenOccupancy, which is the first benchmark for surrounding semantic occupancy perception in driving scenarios. Specifically, we introduce the nuScenes-Occupancy, which extends the nuScenes dataset with dense semantic occupancy annotations based on the proposed AAP pipeline. In the OpenOccupancy benchmark, we establish camera-based, LiDAR-based and multi-modal baselines. Additionally, the CONet is proposed to alleviate the computational burden of high-resolution occupancy predictions. Comprehensive experiments are conducted on the OpenOccupancy benchmark, where the results show that camera-based and LiDAR-based baseline are complementary to each other, and multi-modal baseline further enhances the performance by 47\% and 29\%.
Besides, the proposed CONet relatively improves the baseline by $\sim$30\% with minimal latency overhead. We hope the OpenOccupancy benchmark will be beneficial in the development of surrounding semantic occupancy perception.

{\small
\bibliographystyle{ieee_fullname}
\bibliography{egbib}
}

\end{document}